\documentclass[a4paper,conference]{IEEEtran}
\usepackage{cite}

\ifCLASSINFOpdf
   \usepackage[pdftex]{graphicx}
   \graphicspath{{../pdf/}{../jpeg/}}
   \DeclareGraphicsExtensions{.pdf,.jpeg,.png}
\else
\fi
\usepackage{amsmath}
\usepackage{amssymb}
\usepackage{breqn}
\usepackage{bm}
\usepackage{algorithmic}

\usepackage{array}
\usepackage{multirow}
\newcolumntype{V}{>{\setbox0=\hbox\bgroup}c<{\egroup}@{\hspace*{-\tabcolsep}}}
\newcolumntype{C}[1]{>{\centering\let\newline\\\arraybackslash\hspace{0pt}}m{#1}}
\usepackage{url}

\usepackage[dvipsnames]{xcolor}
\begin{document}

\title{Look Beyond Bias with Entropic Adversarial Data Augmentation}

\author{\IEEEauthorblockN{Thomas Duboudin $^1$}
\and
\IEEEauthorblockN{Emmanuel Dellandréa $^1$}
\and
\IEEEauthorblockN{Corentin Abgrall $^2$}
\and
\IEEEauthorblockN{Gilles Hénaff $^2$}
\and 
\IEEEauthorblockN{Liming Chen $^1$}
\and
\IEEEauthorblockA{\small $^1$ Univ Lyon, Ecole Centrale de Lyon, CNRS, INSA Lyon,\\ Univ Claude Bernard Lyon 1, Univ Louis Lumière Lyon 2,\\ LIRIS, UMR5205, 69134 Ecully, France \\
{\{thomas.duboudin, emmanuel.dellandrea, liming.chen\}@ec-lyon.fr}}
\and
\IEEEauthorblockA{\small $^2$ Thales LAS France \\
{\{corentin.abgrall, gilles.henaff\}@fr.thalesgroup.com}}
}


%


\maketitle


\begin{abstract}




Deep neural networks do not discriminate between spurious and causal patterns, and will only learn the most predictive ones while ignoring the others. This shortcut learning behaviour is detrimental to a network's ability to generalize to an unknown test-time distribution in which the spurious correlations do not hold anymore. Debiasing methods were developed to make networks robust to such spurious biases but require to know in advance if a dataset is biased and make heavy use of minority counter-examples that do not display the majority bias of their class. In this paper, we argue that such samples should not be necessarily needed because the "hidden" causal information is often also contained in biased images. To study this idea, we propose 3 publicly released synthetic classification benchmarks, exhibiting predictive classification shortcuts, each of a different and challenging nature, without any minority samples acting as counter-examples. First, we investigate the effectiveness of several state-of-the-art strategies on our benchmarks and show that they do not yield satisfying results on them. Then, we propose an architecture able to succeed on our benchmarks, despite their unusual properties, using an entropic adversarial data augmentation training scheme. An encoder-decoder architecture is tasked to produce images that are not recognized by a classifier, by maximizing the conditional entropy of its outputs, and keep as much as possible of the initial content. A precise control of the information destroyed, via a disentangling process, enables us to remove the shortcut and leave everything else intact. Furthermore, results competitive with the state-of-the-art on the BAR dataset ensure the applicability of our method in real-life situations.

\end{abstract}


%

\section{Introduction}

Deep neural networks are now the preferred method in computer vision when facing classification, object detection, or semantic segmentation tasks, as they exhibit human-level performances on such kind of tasks \cite{he2016deep, he2017mask}. However, a mismatch between the training and testing data distribution often leads to a sharp drop in performance \cite{hoffman2018cycada}. One of the reasons behind the inability of deep networks to ensure good performance on unseen data distribution is the frequent presence of biases in the training dataset \cite{singla2021understanding, hermann2020shapes, tommasi2015deeper, fabbrizzi2021survey}. If these biases are the most predictive patterns for the task, a network will learn to use them, and ignore the less effective ones, to make its decision (a behaviour called shortcut learning \cite{dagaev2021toogoodtobetrue}, a bias is a spurious shortcut). On a new data distribution, with different biases, the network will not be able to use the previously learned patterns to take proper decisions.






Several strategies have been designed to prevent deep networks from overly focusing on the biases. A first family of works assumes that the bias is given as an auxiliary label (such as in \cite{Kim_2019_CVPR}) on the images, other works \cite{nam2020learning, dagaev2021toogoodtobetrue, ahmed2021systematic, lee2021learning, liu2021just, kim2021biaswap} consider that the bias is what is naturally learned by a model. Both aims to make the network invariant to the bias. The second family of works still requires to know that the dataset is strongly biased and that what is naturally learned can be safely ignored, without losing information. Most of these debiasing methods rely on particular training samples that do not exhibit the majority bias and increase their importance in the training procedure to prevent the model from learning the biases.\\

\begin{figure}[t!]
\centering
\includegraphics[width=3.5in]{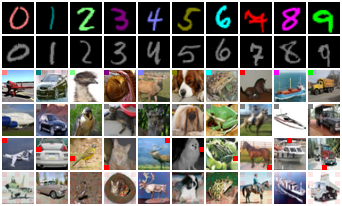}
\caption{Samples of our benchmarks. First and second rows are our Colored-MNIST training and test data. Third and fourth rows are training and test data for the Colored-Patch CIFAR10 benchmark, fifth and last rows are the training and test data for the Located-Patch CIFAR10 benchmark}
\label{samples}
\end{figure}

\begin{figure*}[t!]
\centering
\includegraphics[width=0.9\textwidth]{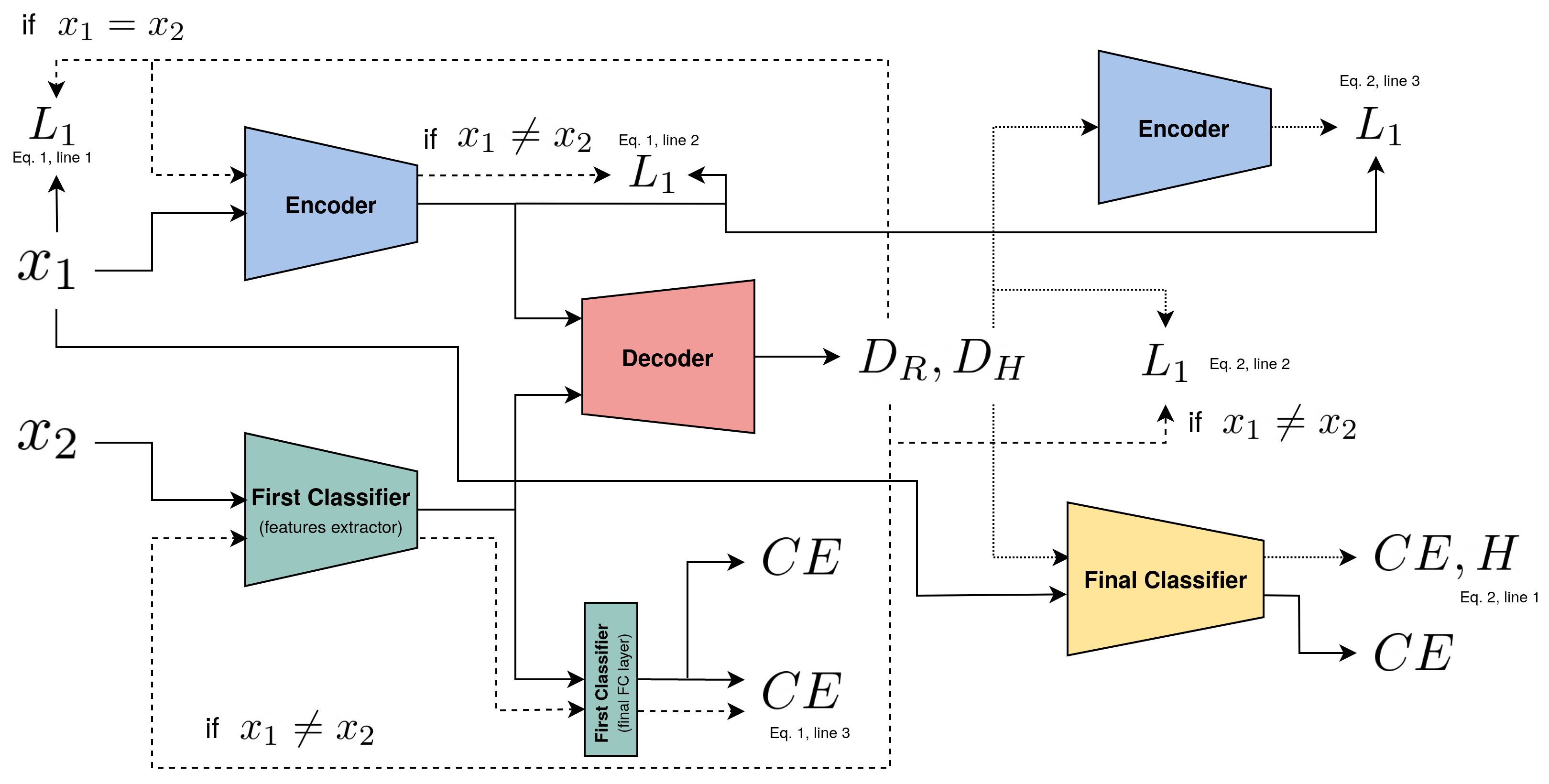}
\caption{Overview of our architecture. Full lines denote the algorithm pipeline for the original images, dashed lines the pipeline for the reconstructed images and the dotted lines for the entropic images. $CE$ denotes a cross-entropy loss, $H$ the conditional entropy of the final classifier, and $L_1$ an $L_1$ distance loss. The 2 encoders displayed refer to the unique one used, but were displayed twice for readability.}
\label{archi}
\end{figure*}

We argue that it is desirable to design methods that do not rely on such few portion of samples. In real-life situations, it could lead to overestimate the importance of outliers that should be ignored, such as annotation errors (Northcutt \textit{et al.} \cite{northcutt2021pervasive} estimated that the average annotation errors percentage in usual computer vision datasets was $3.3\%$). It may also be optimistic to find counter-examples of the biases in some situations, \textit{e.g.} with synthetic datasets where there is often a low diversity of situations. Furthermore, most of the time, the causal patterns are still present in the biased images, they are simply "hidden" behind the more effective biases. Finding the causal patterns in the biased samples should therefore be possible, and probably leads to better generalization than counter-examples based debiasing methods due to the reliance on a vastly larger amount of data. Besides, knowing in advance that a dataset is biased is often an unrealistic hypothesis (spurious biases can be noticeable background or context, as in \cite{beery2018recognition}, but also inconspicuous hospital specific clues, as in \cite{degrave2021ai}). \\

To study debiasing without the help of minority samples, we first design 3 synthetic benchmark datasets, for a classification task, based on the CIFAR-10 \cite{krizhevsky2009learning} and MNIST \cite{mnist} datasets, without any counter-examples. The nature of the biases in the datasets is crafted to be diverse and challenging. On our Colored-MNIST, the bias is spatially mingled with the underlying causal patterns, on our Colored-Patch CIFAR10, the bias is a very local one, and for our Located-Patch CIFAR10, the biases are not texture-based biases but positional ones. Samples of our dataset can be found in Figure \ref{samples}. Experiments on state-of-the-art strategies show that the benchmarks are indeed challenging since no methods yield satisfying results on them. Furthermore, to solve the shortcut issue without using particular data samples, we draw inspiration from methods coming from the domain generalization research field, in which the goal is to make networks robust to unknown domain shifts. From a biased image, we aim to create an image from which the shortcut, or bias, is "removed", but for which everything else is kept intact. Using an encoder-decoder architecture, we transform biased images into images that maximize the entropy of a classifier (trained on both original and transformed images). With a precise control of the amount of information destroyed, via a disentanglement process, we are able to generate images where only the shortcuts are noticeably altered and replaced by patterns that not recognized by the classifier. A classifier trained on these transformed images learn the previously missed patterns since the most obvious ones are now missing. Since we do not assume that the shortcuts were spurious (in a real-life situation, what is naturally learned by a network will be a complex mix of causal and spurious features), we train a classifier on both original and transformed images to learn both the shortcuts and the less effective patterns.\\

To summarize, our contributions are threefold :

\begin{itemize}
    \item We design 3 synthetic biased benchmark datasets, for a classification task, based on the CIFAR10 and MNIST datasets. 
    \item We show that existing works only very partially mitigate the accuracy drop at test-time on our datasets.
    \item We propose a novel method relying on an adversarial encoder-decoder model that removes the bias from the images via entropy maximization. We demonstrate the effectiveness of our strategy on our datasets and on the more realistic BAR \cite{nam2020learning} dataset.
\end{itemize}

\section{Related Works}

\subsection{Debiasing}

Deep neural networks learn correlations between patterns and labels without any regard to how spurious they are \cite{nam2020learning, Kim_2019_CVPR}. In a situation where a dataset is heavily biased, this behaviour will hinder generalization but this issue might be mitigated by having the network ignore the biased patterns even though they are predictive. Kim \textit{et al.} \cite{Kim_2019_CVPR}, are able to make a network invariant to biased patterns, but require the bias to be given as an auxiliary label. This is tedious from a human annotation perspective and restrict the application of the method to a bias that can be spotted with the human eye. Based on this limitation, another line of work \cite{nam2020learning, dagaev2021toogoodtobetrue, ahmed2021systematic, lee2021learning, liu2021just, kim2021biaswap} focuses on finding counter-examples \textit{i.e.}, samples that do not share the majority bias of their class, often via a loss-based criterion. Once such samples are found, their importance during the training is increased via an over-sampling scheme as in Just-Train-Twice (JTT) \cite{liu2021just} or via a loss re-weighting scheme as in Learning-from-Failure (LfF) \cite{nam2020learning}. Closest to our work is the work of Kim \textit{et al.} \cite{kim2021biaswap} that allows the creation of a debiased dataset from a biased one, by generating hybrid images in which biases and labels are decorrelated. A model is then simply trained on this decorrelated dataset and does not learn the biases as they are no longer predictive. Iterating over this idea, in \cite{lee2021learning} (LDD), Lee \textit{et al.} propose to train on "virtual" hybrids by disentangling bias and causal patterns at features level and creating combinations unseen in the training set before training on them. 


\subsection{Domain Generalization}

Domain generalization is a research field aiming to make deep networks robust to unknown and unforeseen domain shifts between training and test-time, without having any information about the test-time domain (such as non-annotated samples from the target domain, as in unsupervised domain adaptation \cite{ganin2015unsupervised, tzeng2017adversarial}). Most algorithms assume to have access to data coming from several identified different domains and aim to learn more robust features by learning features that are shared among all source domains \cite{tzeng2017adversarial, carlucci2019hallucinating, li2018domain}. In single-source domain generalization, however, only one domain is available at training time. In such situation, it is no longer possible to find domain-invariant features, and some methods rely instead on finding more diverse and semantically different patterns than normal. Carlucci \textit{et al.} \cite{carlucci2019domain} (Jigsaw), for instance, used a self-supervised objective alongside the classification, based on solving jigsaw puzzles: having an auxiliary task not related to classification enables the network to learn less domain specific patterns. Representation Self-Challenging (RSC) \cite{huangRSC2020} is a dropout strategy in which the muted coefficients in the intermediary features are the ones most responsible for the prediction and not randomly chosen ones. This therefore forces the network to use less strongly correlated patterns. Spectral Decoupling, a method introduced by Pezeshki \textit{et al.} \cite{pezeshki2020gradient}, proposes an $L_2$ regularization on the output logits of the network (instead of the weights), to combat the "gradient starvation" phenomenon and prevent the network from learning only a subset of the useful patterns. Finally, adversarial data augmentation \cite{ada} has been used to generate samples outside of the training distribution by maximizing the cross-entropy classification loss. In this context, Zhao \textit{et al.} \cite{zhaoNIPS20maximum} used entropy maximization as a secondary regularizer to further push samples away from the training manifold. Our method only uses entropy maximization and a strict control mechanism to only alter the shortcuts.

\subsection{Disentanglement}


Disentanglement is extensively used in the image-to-image translation field where the content of an image is often divided between domain-specific scene-invariant and scene-specific domain-invariant information \cite{huang2018multimodal, gonzalez2018image, swappingautoencoder}. Our method is inspired by image-to-image translation strategies, such as MUNIT \cite{huang2018multimodal}, but applied to a different content division. We disentangle the content into semantic (what is learned by a network solving a task), and non-semantic (the remaining information contained in the image), which may contain information useful for the task but only what is normally overlooked by a network.

\section{Benchmark Datasets}

We introduce 3 benchmark datasets based on the MNIST, and CIFAR-10 datasets. The datasets are publicly available on github \footnote{https://github.com/liris-tduboudin/Look-Beyond-Bias}.

\subsection{Colored-MNIST}

In the training set of this biased dataset, every digit is colored with its particular class color. All the colors chosen are fairly different, and there are no counter-examples. The validation dataset is colored with the same colors as the training set. In the test set however, all the digits are colored with the same color no matter the class so that the accuracy on the test set reflects how well the network learned the less efficient patterns, \textit{i.e.} the shapes of the digits. The test color is the average of all training colors so that the network can't rely on color anymore to classify the images. Other works have introduced color-biased MNIST datasets \cite{ahmed2021systematic,pezeshki2020gradient, arjovsky2020invariant}. Ours differs from theirs most notably by the lack of counter-examples, but also by having a single domain for training, and by having an unbiased dataset for test rather than a dataset differently biased \textit{e.g.} with different class-color combinations. One particularity of the colored MNIST datasets is the fact that the shape and color information are spatially mingled, which prevents a simple training with cropping or cutout \cite{cutout} data augmentation to find the shape information.

\subsection{Colored-Patch CIFAR10}

Inspired by the work of \cite{dagaev2021toogoodtobetrue}, we also design a more complex benchmark dataset based on CIFAR10. For each of the training images, a 5x5 pixels colored patch is added in the top left corner. The color of the patch is the image's class color, again with no counter-examples. In the test set all the images have the same patch color, which is the average of the training colors. For this dataset, the bias is spatially located in a tiny part of the images instead of being global as with the MNIST dataset. The colors used are the same than the ones used for the previous dataset.

\subsection{Located-Patch CIFAR10}

Finally, to experiment on a biased dataset for which the bias is not texture-based, we create a CIFAR10-based benchmark where it is the position of a 5x5 pixels patch that is absolutely correlated with the label \textit{e.g.} top left corner patch for planes, bottom right corner for horses. The color of the added patch is red, for all classes. For the test dataset, the average patch is added to all the images no matter their class.

\section{Entropic Adversarial Data Augmentation}

\subsection{Proposed Method}

Our method uses 4 distinctly trained neural networks: 2 classifiers: a first classifier $C_0$, with its convolutional features extractor $F_0$, and a final classifier $C_f$, with the same architecture, one encoder $E$, and one decoder $D$. The decoder $D$ takes as input the output of the encoder $E(x)$, and the output of the features extractor $F_0(x)$. These 2 feature maps are simply resized and concatenated in the channel dimension before being sent to the decoder. We chose this fusion strategy against AdaIN \cite{huang2017arbitrary} based methods, such as \cite{kim2021biaswap}, because it is simpler and makes it easier for the decoder to learn spatial information. The decoder outputs simultaneously 2 images (the output image has 6 channels, the first 3 being the reconstructed image $D_R$, and the remaining 3 the entropic images image $D_H$). \\


The first classifier $C_0$ is trained to minimize the cross-entropy on the original images, without alteration to a standard training procedure. The encoder, through a latent space reconstruction loss, is made invariant to the shortcuts learned by the first classifier $C_0$. The decoder $D_R$ is conditioned to output the encoder's input content with the features extractor's $F_0$ input shortcut. This architecture aims to produce a disentangled representation of an image between the features used by the first classifier $C_0$ and the remaining information $E(x)$ needed to reconstruct the original images. The remaining 3 channels of the decoder ($D_H$) are used to generate the entropic images via an adversarial training scheme. Samples of such hybrid and entropic images can found in Figure \ref{generated_samples}. The final classifier is simply trained to classify both the original and the entropic images coming from $D_H$ to learn both the shortcuts (not necessarily spurious) and the "hidden" patterns. All the networks are simultaneously trained with their corresponding losses, though the first classifier can be trained offline beforehand and frozen during the training of the other networks. A full schema of the proposed method is available in Figure \ref{archi}.



\begin{align}
    \nonumber &\mathcal{L}_R(D_R, E) = \alpha \mathbb{E}_{x\sim p_{x}}[||D_R(E(x),F_0(x))-x||_1] \\
    \nonumber &+ \beta \mathbb{E}_{(x_1,x_2)\sim {p_x}^2}[||E(D_R(E(x_1),F_0(x_2)))-E(x_1)||_1] \\
    \nonumber &+ \gamma \mathbb{E}_{(x_1,x_2)\sim {p_x}^2}[-\sum_i \delta_i(C_0(x_2)) \times \\
    &\text{log}\delta_i(C_0(D_R(E(x_1),F_0(x_2)))) ]
    \label{reconstruction_optim}
\end{align}

To properly condition the encoder and the decoder $D_R$ to yield a disentangled representation, several training objectives are required: a reconstruction loss in the image space between the original image and the reconstructed one (Eq.\ref{reconstruction_optim}, line 1), an encoder latent space reconstruction loss (Eq.\ref{reconstruction_optim}, line 2), and a classifier prediction consistency loss (Eq.\ref{reconstruction_optim}, line 3). The classifier consistency loss is a cross-entropy between the prediction on an original image $x_2$, $C_0(x_2)$, and the prediction on a hybrid image created from $E(x_1)$ and $F_0(x_2)$: $C_0(D_R(E(x_1), F_0(x_2)))$, with $\delta_i$ being the $i$-th softmax coefficient: $\delta_i(y) = e^{y_i} / \sum_j e^{y_j}$. The last 2 losses require the sampling of 2 images simultaneously ($x_2$ can be obtained by applying a permutation on the current batch along the sample dimension). These losses enable the encoder to learn all necessary patterns for the reconstruction task but the shortcuts already provided by the features extractor $F_0$, and prevent the decoder from inferring the shortcuts from the encoder representation, as the shortcut of its hybrid output $D_R(E(x_1), F_0(x_2))$ must be the one of $x_2$. The first and the final classifiers are not optimized with regards to these constraints. 

\begin{align}
    \nonumber &\mathcal{L}_H(D_H) = \varepsilon \mathbb{E}_{x\sim p_x}[- \sum_i{q\text{log}\delta_i(C_f(D_H(E(x), F_0(x))))}] \\
    \nonumber &+ \mu \mathbb{E}_{x\sim p_x}[||D_H(E(x),F_0(x)) - \mathbb{E}_{f_0 \sim p_{f_0}}[D_H(E(x), f_0)||_1] \\
    &+ \nu \mathbb{E}_{x \sim p_x}[||E(D_H(E(x),F_0(x))) - E(x)||_1]
    \label{entropic_optim}
\end{align}

The entropic output of the decoder $D_H$ is trained to maximize the entropy of the final classifier, that is trained on both the original images and the entropic ones. The rational behind this adversarial loss is that high-entropy images do not contain patterns that can be preferentially linked to a class, and hence are more likely to be devoid of the original shortcuts. Maximizing the entropy of the first classifier only leads to changes in the bias, and not to the complete removal we are aiming for. Alongside the entropy maximization (Eq. \ref{entropic_optim}, line 1, with $q$ the uniform probability density: $q=1/N_c$ where $N_c$ is the number of classes), the entropic images are subject to several constraints to avoid the destruction of all information. The first constraint lies in the decoder itself: up until the last layer, the weights are shared for both the entropic images generation and the reconstruction task. We also use the encoder latent space reconstruction loss (Eq.\ref{entropic_optim}, line 3) on the ground that the entropic images should precisely not modify what is extracted by the encoder (everything but the shortcut). Finally, we use an encoder-conditioned expected image reconstruction loss (Eq.\ref{entropic_optim}, line 2). This loss aims to drive the entropic images toward an image that should already be confusing for the classifier, while keeping the information not used in classification intact. It is implemented by minimizing the $L_1$ loss between the entropic image $D_H(E(x_1),F_0(x_1))$ and an hybrid image generated from the encoding of $x_1$: $D_R(E(x_1),F_0(x_2))$. Because the semantic image $x_2$ is randomly sampled every iteration, minimizing this loss will eventually yield the average-biased image $\mathbb{E}_{f_0\sim p_{f_0}}[D_R(E(x_1), f_0)]$. 


\subsection{Baselines for comparison}

We compare ourselves with several state-of-the-art strategies from either the debiasing or the domain generalization community. A change of biases between training and test is a domain shift, and even though domain generalization methods were not developed with such shift in mind, it is interesting to see how they perform. We also compare ourselves with simple baselines as a reality check: first, with the standard training procedure (stochastic gradient descent with momentum, with the cross-entropy loss), then, with dropout \cite{dropout}. Dropout is a naive way of finding more useful patterns, by preventing the network to use all the available information at disposal. Since nothing prevents the network from learning the same patterns in several filters, we experiment with an orthogonality constraints on the weights (inspired by \cite{ortho}) to force neurons to be different, and with a constraint over the covariance matrix of the intermediary activations (inspired by \cite{correlation}), on the ground that filters that activate very often together are likely to check for the same patterns. Finally, we compare ourselves with the single-source domain generalization methods (Jigsaw, Spectral Decoupling, and RSC) and debiasing methods (LfF, JTT and LDD) reviewed in the related works section. 


\section{Experiments and Discussion}

\subsection{Experimental setup}


Our architecture exists in 2 different flavors : large-scale, and small-scale. The large-scale version uses a ResNet18 \cite{he2016deep} (adapted to the CIFAR10 and MNIST datasets as in \cite{yun2019cutmix}) as classifier, and a UNet \cite{ronneberger2015u} as encoder and decoder. The features extractor $F_0$ is the classifier without its last layer. The UNet is divided into a multi-output encoder and a multi-input decoder to account for the skip-connections. Each encoder output is taken as input by the corresponding decoder input. The semantic information from the features extractor $F_0$ is concatenated to the deepest encoder output before being given to the deepest decoder input. The small-scale version uses a LeNet as classifier (as used in \cite{carlucci2019domain} for the MNIST-based experiments), a 4-layers convolutional network as encoder, and a 4-layers decoder with transposed convolutions. For both architecture, the optimizer used is Adam \cite{kingma2015adam} with the same learning rates for all the networks. We evaluate our approach on our synthetic benchmarks and on the more realistic Biased-Action-Recognition (BAR) dataset. BAR images are divided into 6 activity classes and exhibit a strong (but not absolute) background bias in the training data: climbing often takes place in on grey rocky background, throwing on a green baseball field, \textit{etc}. The test set however is mostly made of images taken from unusual circumstances. This dataset is a common benchmark for debiasing methods that make use of the training minority samples. No data augmentation is used for the experiments as we want to assess the effectiveness of our method without adding any predefined invariances to the training procedure. Hyper-parameters settings and model selection are done without using the test set, to mimic a situation where the test-time data distribution is unknown. For our architecture, we use the entropy loss curve to set the sensitive $\varepsilon$ hyper-parameters: it should, in fact, increase continuously during training. A diminishing entropy means that the final classifier cannot keep up with the decoder's entropic images and that there is a destruction of information. The goal is to aim for the slowest increasing entropy curve possible. For the model selection with our method, we use the model at the final epoch, on the ground that training on the entropic images is much slower than the training on the normal images. Training is considered complete when the entropy curve is no longer increases. Finally, because the effect of the random initialization of the network is greater than usual in a domain shift situation \cite{krueger2020out}, results are averaged over 3 runs. More details about the architecture and the training hyper-parameters are available in the supplementary material.

\subsection{Results and analysis}

\begin{figure}[h!]
\centering
\includegraphics[width=8.8cm]{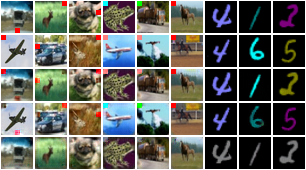}
\caption{Samples of generated images. First \& second rows: original images. Third row: hybrid images with first row content and second row shortcuts. Fourth row: hybrid images with second row content and first row shortcuts. Last row: entropic images of the first row.}
\label{generated_samples}
\end{figure}

\begin{table*}[h!]
\scriptsize
\caption{Main results}
\label{main_table}
\begin{center}
\begin{tabular}{|l||c|c|c|c|c|c|c|c| V|}
\hline
\multicolumn{9}{|c|}{\textbf{Large-Scale Experiments (Resnet18 as classifier)}}\\
\hline
Dataset $\rightarrow$ & \multicolumn{2}{c|}{Colored-MNIST} & \multicolumn{2}{c|}{Colored-Patch CIFAR10} & \multicolumn{2}{c|}{Located-Patch CIFAR10} & \multicolumn{2}{c|}{BAR}\\
\hline
Method $\downarrow$ & Val. Acc. & Test Acc. & Val. Acc. & Test Acc. & Val. Acc. & Test Acc. & Val. Acc. & Test Acc.\\
\hline\hline
Standard Training Procedure & $100 \pm 0.0$ & $18.8 \pm 7.4$& $100 \pm 0.0$ & $10.3 \pm 0.5$ & $100 \pm 0.0$ & $10.0 \pm 0.0$ & $97.5 \pm 0.6$ & $49.1 \pm 1.9$ \\
\hline
Dropout & $100 \pm 0.0$ & $14.6 \pm 6.2$ & $100 \pm 0.0$ & $11.0 \pm 1.8$ & $100 \pm 0.0$ & $10.1 \pm 0.1$ & $98.9 \pm 0.3$ & $49.4 \pm 1.0$ \\
\hline
Dropout \& Orthogonality \cite{ortho} & $100 \pm 0.0$ & $10.2 \pm 0.1$ & $100 \pm 0.0$ & $10.0 \pm 0.0$ & $100 \pm 0.0$ & $10.0 \pm 0.0$ & $98.8 \pm 0.6$ & $48.0 \pm 1.4$\\
\hline
Dropout \& Covariance \cite{correlation} & $93.3 \pm 3.8$ & $27.2 \pm 5.4$ & $86.4 \pm 10.5$ & $21.6 \pm 2.4$ & $67.9 \pm 9.5$ & $26.3 \pm 2.2$ & $90.6 \pm 3.1$ & $28.0 \pm 3.2$\\
\hline
Jigsaw Puzzle \cite{carlucci2019domain} & $99.8 \pm 0.3$ & $21.5 \pm 4.9$ & $99.9 \pm 0.0$ & $17.8 \pm 0.9$ & $100 \pm 0.0$ & $12.1 \pm 0.5$ & $97.5 \pm 0.3$ & $49.8 \pm 1.8$\\
\hline
Spectral Decoupling \cite{pezeshki2020gradient} & $99.9 \pm 0.0$ & $24.5 \pm 2.9$ & $100 \pm 0.0$ & $10.4 \pm 0.35$ & $100 \pm 0.0$ & $10.2 \pm 0.1$ & $96.1 \pm 0.6$ & $44.5 \pm 1.8$ \\
\hline
RSC \cite{huangRSC2020} & $100 \pm 0.0$ & $12.5 \pm 1.9$ & $96.7 \pm 5.7$ & $10.0 \pm 0.1$ & $100 \pm 0.0$ & $10.0 \pm 0.0$ & $96.8 \pm 0.9$ & $50.2 \pm 1.4$ \\
\hline
LfF \cite{nam2020learning} & $100 \pm 0.0$ & $9.8 \pm 1.9$ & $100 \pm 0.0$ & $10.2 \pm 0.3$ & $100 \pm 0.0$ & $10.6 \pm 0.7$ & $97.4 \pm 0.3$ & $54.3 \pm 2.3$ \\
\hline
JTT \cite{liu2021just} & $100 \pm 0.0$ & $12.5 \pm 4.3$ & $100 \pm 0.0$  &$10.4 \pm 0.4$ & $100 \pm 0.0$ & $10.0 \pm 0.02$ & $97.3 \pm 0.8$ & $50.2 \pm 2.9$ \\
\hline
LDD \cite{lee2021learning} &  $100 \pm 0.0$ & $14.8 \pm 5.3$ & $100 \pm 0.0$ & $10.0 \pm 0.2$ &  $100 \pm 0.0$ & $10.2 \pm 0.3$ & $98.3 \pm 0.8$ & $53.61 \pm 2.7$ \\
\hline
\textbf{Ours} & $\mathbf{99.8 \pm 0.2}$ & $\mathbf{97.3 \pm 0.84}$ & $\mathbf{93.8 \pm 1.3}$ & $\mathbf{78.9 \pm 0.34}$ & $\mathbf{98.0 \pm 0.6}$ & $\mathbf{75.6 \pm 3.8}$ & $\mathbf{97.1 \pm 0.8}$ & $\mathbf{54.4 \pm 1.1}$\\
\hline
Standard Training Procedure & $99.4 \pm 0.04$ & $99.4 \pm 0.04$ & $77.4 \pm 0.08$ & $77.4 \pm 0.08$ & $77.4 \pm 0.08$ & $77.4 \pm 0.08$ & - & - \\
on the original datasets & & & & & & & & \\
\hline
\end{tabular}
\end{center}
\end{table*}



Our main results are available in Table \ref{main_table} for the large-scale version. Small-scale results are available in the supplementary material. The numbers displayed are the average accuracy $\pm$ the standard deviation. Our method yields a significant accuracy improvement over the previous works on our synthetic benchmarks on the test sets, while retaining a very high accuracy on the validation sets. A high accuracy reached in both validation and test indicates that both the shortcuts and the "hidden" patterns are learned: if the shortcuts are completely ignored, we expect the accuracy to be similar for the biased and unbiased datasets. The drop in validation for Colored-Patch CIFAR10 is most likely due to non-optimal hyper-parameters: we used the same hyper-parameters for all 3 synthetic datasets. Final accuracy seems to be moderately sensitive to loss weights, except for the entropy maximization weight. Other strategies perform only marginally better than the standard training procedure. It is not surprising for debiasing methods that require explicit counter-examples. Furthermore, our architecture enables the final classifier to find all the possible patterns "hidden" behind the shortcut: the accuracy of our method on the test dataset is roughly equal to the accuracy we get when training the same network normally on the original MNIST or CIFAR10 datasets (without biases). On the BAR dataset, our method performs on par with the state-of-the-art debiasing methods, without explicitly relying on the unusual samples. Discrepancies between original LfF results and ours is mostly due to the resizing of the images for computational convenience (128x128 in our experiments, 224x224 in the original ones). Samples of our entropic images can found in Figure \ref{generated_samples} and are effectively devoid of the original shortcuts. \\

We also conduct an ablation study of the small-scale version of our architecture on the synthetic benchmark datasets. Our study was conducted to shed the light on 2 questions: 1 - is the disentangling part of the architecture needed \textit{i.e.} can't an entropy maximization and a $L_1$ loss between the transformed image and the original one be sufficient to yield unbiased images ? 2 - is the entropy maximization constraint needed \textit{i.e.} can't the disentangling part with the encoder-conditioned expected image $L_1$ reconstruction loss be enough ? Results of the ablation study are available in Table \ref{ablation_study}. The reported numbers are the average accuracy on the test sets of the datasets. For the first experiment, we conduct a study with varying entropy maximization loss weight $\varepsilon$, the weight for the identity loss $\alpha$ is fixed to 1.0. For the second experiment, all the used weights ($\alpha, \beta, \gamma, \mu$) are fixed to 1, all the others to 0.0. Our study shows that, while either simplified architecture can yield satisfying results on a certain dataset, for a strategy to work well on all benchmarks it has to use all the proposed constraints. Without the disentangling part (Eq. \ref{reconstruction_optim}), the accuracy suffers on all datasets but especially on the Colored-MNIST dataset, no matter the $\varepsilon$ used. We hypothesize that this is due to the spatially widespread bias in the dataset. Without the entropy maximization loss, the architecture is not able to learn anything on the CIFAR10-based benchmarks. For the Located-Patch CIFAR10, this is due to the positional nature of the bias: without the entropy maximization constraint, the patch is simply replaced by a brown patch (similar to what can be seen in Figure \ref{generated_samples} with the hybrid images). A classification model trained on these samples will simply make use of a differently colored patch to take its decisions. There are no constraints to push the encoder-decoder to realistically inpaint the missing patch (such as a co-occurrence discriminator loss \cite{swappingautoencoder}) as the entropy maximization is effective enough. For the Colored-Patch CIFAR10, imperfections in the disentanglement process prevents the decoder from completely removing the original color of the patch.


\begin{table}[h!]
\caption{Ablation study}
\label{ablation_study}
\begin{center}
\begin{tabular}{|C{1.9cm}||C{1.2cm}|C{1.7cm}|C{1.7cm}|}
\hline
Dataset $\rightarrow$ & Colored-MNIST & Colored-Patch CIFAR10 & Located-Patch CIFAR10 \\
\hline\hline
1 - no disentanglement & & & \\
$\varepsilon = 10^{-4}$ & $67.1$ & $63.5$ & $12.1$\\
$\varepsilon = 10^{-3}$ & $63.6$ & $69.3$ & $53.5$ \\
$\varepsilon = 10^{-2}$ & $31.2$ & $62.7$ & $60.3$\\
$\varepsilon = 10^{-1}$ & $49.1$ & $35.7$ & $22.4$ \\
$\varepsilon = 1.0$ & $25.6$ & $14.4$ & $16.0$ \\
\hline
2 - no entropy maximization & $97.7$ & $16.0$ & $17.3$ \\
\hline
Full Method & $98.4$ & $69.2$ & $67.3$ \\
\hline
\end{tabular}
\end{center}
\end{table}

\section{Conclusion}

In this paper, we created 3 different benchmarks to study the behaviour of domain generalization and debiasing methods when facing a dataset where the bias is shared by all the samples without exception. No existing work yield satisfying results on it. We then proposed a generative architecture relying on entropic adversarial data augmentation and on disentangling a representation between shortcuts and remaining useful patterns and showed that it performed as well as possible considering the classifiers used, on our 3 benchmarks. Further experiments on the BAR dataset yielded results competitive with state-of-the-art methods. This is an indication that the explicit search for counter-examples might not be necessarily needed: the information usually overlook by neural networks is contained even in samples that exhibit the shortcuts. The future works will be dedicated to the study of more real-life like situations.





%

\bibliographystyle{IEEEtran}
\bibliography{IEEEabrv, biblio}

\begin{thebibliography}{10}
\providecommand{\url}[1]{#1}
\csname url@samestyle\endcsname
\providecommand{\newblock}{\relax}
\providecommand{\bibinfo}[2]{#2}
\providecommand{\BIBentrySTDinterwordspacing}{\spaceskip=0pt\relax}
\providecommand{\BIBentryALTinterwordstretchfactor}{4}
\providecommand{\BIBentryALTinterwordspacing}{\spaceskip=\fontdimen2\font plus
\BIBentryALTinterwordstretchfactor\fontdimen3\font minus
  \fontdimen4\font\relax}
\providecommand{\BIBforeignlanguage}[2]{{%
\expandafter\ifx\csname l@#1\endcsname\relax
\typeout{** WARNING: IEEEtran.bst: No hyphenation pattern has been}%
\typeout{** loaded for the language `#1'. Using the pattern for}%
\typeout{** the default language instead.}%
\else
\language=\csname l@#1\endcsname
\fi
#2}}
\providecommand{\BIBdecl}{\relax}
\BIBdecl

\bibitem{he2016deep}
K.~He, X.~Zhang, S.~Ren, and J.~Sun, ``Deep residual learning for image
  recognition,'' in \emph{IEEE/CVF Conference on Computer Vision and Pattern
  Recognition}, 2016.

\bibitem{he2017mask}
K.~He, G.~Gkioxari, P.~Doll{\'a}r, and R.~Girshick, ``Mask r-cnn,'' in
  \emph{IEEE/CVF International Conference on Computer Vision}, 2017.

\bibitem{hoffman2018cycada}
J.~Hoffman, E.~Tzeng, T.~Park, J.-Y. Zhu, P.~Isola, K.~Saenko, A.~Efros, and
  T.~Darrell, ``Cycada: Cycle-consistent adversarial domain adaptation,'' in
  \emph{International Conference on Machine Learning}, 2018.

\bibitem{singla2021understanding}
S.~Singla, B.~Nushi, S.~Shah, E.~Kamar, and E.~Horvitz, ``Understanding
  failures of deep networks via robust feature extraction,'' in \emph{IEEE/CVF
  Conference on Computer Vision and Pattern Recognition}, 2021.

\bibitem{hermann2020shapes}
K.~Hermann and A.~Lampinen, ``What shapes feature representations? exploring
  datasets, architectures, and training,'' \emph{Advances in Neural Information
  Processing Systems}, 2020.

\bibitem{tommasi2015deeper}
T.~Tommasi, N.~Patricia, B.~Caputo, and T.~Tuytelaars, ``A deeper look at
  dataset bias,'' in \emph{German Conference on Pattern Recognition}, 2015.

\bibitem{fabbrizzi2021survey}
S.~Fabbrizzi, S.~Papadopoulos, E.~Ntoutsi, and I.~Kompatsiaris, ``A survey on
  bias in visual datasets,'' 2021.

\bibitem{dagaev2021toogoodtobetrue}
N.~Dagaev, B.~D. Roads, X.~Luo, D.~N. Barry, K.~R. Patil, and B.~C. Love, ``A
  too-good-to-be-true prior to reduce shortcut reliance,'' \emph{arXiv preprint
  arXiv:2102.06406}, 2021.

\bibitem{Kim_2019_CVPR}
B.~Kim, H.~Kim, K.~Kim, S.~Kim, and J.~Kim, ``Learning not to learn: Training
  deep neural networks with biased data,'' in \emph{IEEE/CVF Conference on
  Computer Vision and Pattern Recognition}, June 2019.

\bibitem{nam2020learning}
J.~Nam, H.~Cha, S.~Ahn, J.~Lee, and J.~Shin, ``Learning from failure: Training
  debiased classifier from biased classifier,'' in \emph{Advances in Neural
  Information Processing Systems}, 2020.

\bibitem{ahmed2021systematic}
F.~Ahmed, Y.~Bengio, H.~van Seijen, and A.~Courville, ``Systematic
  generalisation with group invariant predictions,'' in \emph{International
  Conference on Learning Representations}, 2021.

\bibitem{lee2021learning}
J.~Lee, E.~Kim, J.~Lee, J.~Lee, and J.~Choo, ``Learning debiased representation
  via disentangled feature augmentation,'' \emph{Advances in Neural Information
  Processing Systems}, 2021.

\bibitem{liu2021just}
E.~Z. Liu, B.~Haghgoo, A.~S. Chen, A.~Raghunathan, P.~W. Koh, S.~Sagawa,
  P.~Liang, and C.~Finn, ``Just train twice: Improving group robustness without
  training group information,'' in \emph{International Conference on Machine
  Learning}, 2021.

\bibitem{kim2021biaswap}
E.~Kim, J.~Lee, and J.~Choo, ``Biaswap: Removing dataset bias with
  bias-tailored swapping augmentation,'' in \emph{IEEE/CVF International
  Conference on Computer Vision}, 2021.

\bibitem{northcutt2021pervasive}
C.~G. Northcutt, A.~Athalye, and J.~Mueller, ``Pervasive label errors in test
  sets destabilize machine learning benchmarks,'' in \emph{Advances in Neural
  Information Processing Systems}, 2021.

\bibitem{beery2018recognition}
S.~Beery, G.~Van~Horn, and P.~Perona, ``Recognition in terra incognita,'' in
  \emph{IEEE/CVF European conference on computer vision}, 2018.

\bibitem{degrave2021ai}
A.~J. DeGrave, J.~D. Janizek, and S.-I. Lee, ``Ai for radiographic covid-19
  detection selects shortcuts over signal,'' \emph{Nature Machine
  Intelligence}, 2021.

\bibitem{krizhevsky2009learning}
A.~Krizhevsky, G.~Hinton \emph{et~al.}, ``Learning multiple layers of features
  from tiny images,'' 2009.

\bibitem{mnist}
Y.~LeCun, C.~Cortes, and C.~Burges, ``Mnist handwritten digit database,''
  \emph{ATT Labs. Available: http://yann.lecun.com/exdb/mnist}, 2010.

\bibitem{ganin2015unsupervised}
Y.~Ganin and V.~Lempitsky, ``Unsupervised domain adaptation by
  backpropagation,'' in \emph{International Conference on Machine Learning},
  2015.

\bibitem{tzeng2017adversarial}
E.~Tzeng, J.~Hoffman, K.~Saenko, and T.~Darrell, ``Adversarial discriminative
  domain adaptation,'' in \emph{IEEE/CVF Conference on Computer Vision and
  Pattern Recognition}, 2017.

\bibitem{carlucci2019hallucinating}
F.~M. Carlucci, P.~Russo, T.~Tommasi, and B.~Caputo, ``Hallucinating agnostic
  images to generalize across domains,'' in \emph{IEEE/CVF International
  Conference on Computer Vision Workshop}, 2019.

\bibitem{li2018domain}
H.~Li, S.~Jialin~Pan, S.~Wang, and A.~C. Kot, ``Domain generalization with
  adversarial feature learning,'' in \emph{IEEE/CVF Conference on Computer
  Vision and Pattern Recognition}, 2018.

\bibitem{carlucci2019domain}
F.~M. Carlucci, A.~D'Innocente, S.~Bucci, B.~Caputo, and T.~Tommasi, ``Domain
  generalization by solving jigsaw puzzles,'' in \emph{IEEE/CVF Conference on
  Computer Vision and Pattern Recognition}, 2019.

\bibitem{huangRSC2020}
Z.~Huang, H.~Wang, E.~P. Xing, and D.~Huang, ``Self-challenging improves
  cross-domain generalization,'' in \emph{IEEE/CVF European Conference on
  Computer Vision}, 2020.

\bibitem{pezeshki2020gradient}
M.~Pezeshki, S.-O. Kaba, Y.~Bengio, A.~Courville, D.~Precup, and G.~Lajoie,
  ``Gradient starvation: A learning proclivity in neural networks,''
  \emph{arXiv preprint arXiv:2011.09468}, 2020.

\bibitem{ada}
R.~Volpi, H.~Namkoong, O.~Sener, J.~C. Duchi, V.~Murino, and S.~Savarese,
  ``Generalizing to unseen domains via adversarial data augmentation,'' in
  \emph{Advances in Neural Information Processing Systems}, 2018.

\bibitem{zhaoNIPS20maximum}
L.~Zhao, T.~Liu, X.~Peng, and D.~Metaxas, ``Maximum-entropy adversarial data
  augmentation for improved generalization and robustness,'' in \emph{Advances
  in Neural Information Processing Systems}, 2020.

\bibitem{huang2018multimodal}
X.~Huang, M.-Y. Liu, S.~Belongie, and J.~Kautz, ``Multimodal unsupervised
  image-to-image translation,'' in \emph{IEEE/CVF European Conference on
  Computer Vision}, 2018.

\bibitem{gonzalez2018image}
A.~Gonzalez-Garcia, J.~van~de Weijer, and Y.~Bengio, ``Image-to-image
  translation for cross-domain disentanglement,'' \emph{Advances in Neural
  Information Processing Systems}, 2018.

\bibitem{swappingautoencoder}
T.~Park, J.-Y. Zhu, O.~Wang, J.~Lu, E.~Shechtman, A.~Efros, and R.~Zhang,
  ``Swapping autoencoder for deep image manipulation,'' in \emph{Advances in
  Neural Information Processing Systems}, 2020.

\bibitem{arjovsky2020invariant}
M.~Arjovsky, L.~Bottou, I.~Gulrajani, and D.~Lopez-Paz, ``Invariant risk
  minimization,'' \emph{arXiv preprint arXiv:1907.02893}, 2020.

\bibitem{cutout}
T.~DeVries and G.~W. Taylor, ``Improved regularization of convolutional neural
  networks with cutout,'' \emph{arXiv preprint arXiv:1708.04552}, 2017.

\bibitem{huang2017arbitrary}
X.~Huang and S.~Belongie, ``Arbitrary style transfer in real-time with adaptive
  instance normalization,'' in \emph{IEEE/CVF International Conference on
  Computer Vision}, 2017.

\bibitem{dropout}
N.~Srivastava, G.~Hinton, A.~Krizhevsky, I.~Sutskever, and R.~Salakhutdinov,
  ``Dropout: A simple way to prevent neural networks from overfitting,''
  \emph{Journal of Machine Learning Research}, 2014.

\bibitem{ortho}
N.~Bansal, X.~Chen, and Z.~Wang, ``Can we gain more from orthogonality
  regularizations in training deep networks?'' in \emph{Advances in Neural
  Information Processing Systems}, 2018.

\bibitem{correlation}
M.~Cogswell, F.~Ahmed, R.~Girshick, L.~Zitnick, and D.~Batra, ``Reducing
  overfitting in deep networks by decorrelating representations,''
  \emph{Internation Conference on Learning Representations}, 2016.

\bibitem{yun2019cutmix}
S.~Yun, D.~Han, S.~J. Oh, S.~Chun, J.~Choe, and Y.~Yoo, ``Cutmix:
  Regularization strategy to train strong classifiers with localizable
  features,'' in \emph{IEEE/CVF International Conference on Computer Vision},
  2019.

\bibitem{ronneberger2015u}
O.~Ronneberger, P.~Fischer, and T.~Brox, ``U-net: Convolutional networks for
  biomedical image segmentation,'' in \emph{International Conference on Medical
  image computing and computer-assisted intervention}, 2015.

\bibitem{kingma2015adam}
D.~P. Kingma and J.~Ba, ``Adam: A method for stochastic optimization,'' in
  \emph{International Conference on Learning Representations}, 2015.

\bibitem{krueger2020out}
D.~Krueger, E.~Caballero, J.-H. Jacobsen, A.~Zhang, J.~Binas, R.~L. Priol, and
  A.~Courville, ``Out-of-distribution generalization via risk extrapolation
  (rex),'' \emph{arXiv preprint arXiv:2003.00688}, 2020.

\end{thebibliography}

\newpage

\section*{Acknowledgement}

This work was in part supported by the 4D Vision project funded by the Partner University Fund (PUF), a FACE program, as well as the French Research Agency, l’Agence Nationale de Recherche (ANR), through the projects Learn Real (ANR-18-CHR3-0002-01), Chiron (ANR-20-IADJ-0001-01), Aristotle (ANR-21-FAI1-0009-01), and the joint support of the French national program of investment of the future and the regions through the PSPC FAIR Waste project.

\section*{Supplementary Material}

This supplementary material presents additional results and details about the experiments conducted in the paper, that were not directly included due to the 6 pages limit. 

\subsection{Debiasing and label noise}

We want to study the behaviour of debiasing methods when they are applied to a dataset containing annotation errors, or label noise, as their impact might be a potential drawback. To do so, we create another synthetic dataset based on MNIST. The training images are colorized as in our original Colored-MNIST dataset (every images of a particular class are colored the same), but the image label is replaced by a random one (chosen uniformly among all labels) with a certain probability $p$. We used $p=0.01$, which gives $1\%$ of randomly labeled samples. This is the order of magnitude of label noise that is encountered in usual datasets \cite{northcutt2021pervasive}. The test set is from the same data distribution, with label noise. There is no domain shift in this situation. We conducted this experiment with a ResNet18 as classifier for the debiasing methods, and with the large-scale version of our architecture. Results of debiasing methods and of our approach on this dataset are available in Table \ref{annotation_errors_table}.

\begin{table}[h!]
\renewcommand{\arraystretch}{1.3}
\caption{Debiasing and label noise}
\label{annotation_errors_table}
\centering
\begin{tabular}{|c||c|}
\hline
Method & Colored-MNIST w. label noise\\
\hline
LDD \cite{lee2021learning}  & $99.1$ \\
\hline
JTT \cite{liu2021just} & $99.3$ \\
\hline
LfF \cite{nam2020learning} & $63.9$ \\
\hline
Standard Training & $99.2$ \\
\hline
Ours & $99.0$ \\
\hline
\end{tabular}
\end{table}

\begin{figure}[h!]
\centering
\includegraphics[width=8cm]{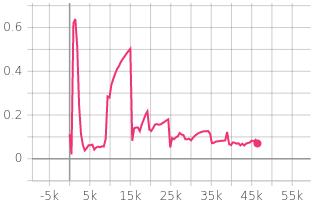}
\caption{Test accuracy over training iterations for LfF, on our Colored-MNIST with label noise.}
\label{annotation_errors_fig}
\end{figure}

The only method that does not succeed in dealing with the label noise is LfF: training collapses after a few iterations (see Figure \ref{annotation_errors_fig}), and does not recover. The best test accuracy is reached at the very beginning of the training, before the minority samples are noticeably over-weighted, and even then it is far below the other works. All the other methods yield perfect results. Training debiasing methods on a dataset with wrongly labeled samples might have a adverse effect on the resulting accuracy, depending on the precise strategy used.

\subsection{Ablation study}

\begin{figure*}[t!]
\centering
\includegraphics[width=18cm]{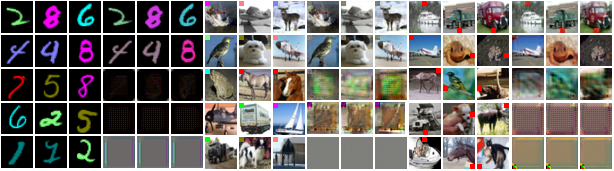}
\caption{Samples of generated entropic images for the first ablation study. The first row corresponds to an entropy maximization objective weight of $\varepsilon=10^{-4}$. This weight is increased by a factor 10 between each row. For every dataset, the first 3 columns are the original images, and the remaining 3 the corresponding entropic images.}
\label{generated_samples_entropic}
\end{figure*}

\noindent
\textbf{1} - for the first ablation experiment, the reconstruction output of the decoder $D_R$ is no longer trained with any objectives. The entropic output $D_H$ and the encoder $E$ are trained with both an entropy maximization and an image reconstruction loss, between the entropic image and the original one. Samples of generated entropic images for this ablation are available in Figure \ref{generated_samples_entropic}. The visualization of the samples confirms the quantitative results: too much information starts to be destroyed for $\varepsilon=10^{-2}$, and too little for $\varepsilon=10^{-4}$, hence the drops in test-time accuracy at both ends of the weight range for Located-Patch CIFAR10, and Colored-MNIST.

\begin{figure}[h!]
\centering
\includegraphics[width=8.2cm]{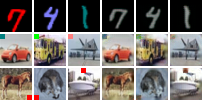}
\caption{Samples of generated encoder-conditioned expected images for the second ablation study.}
\label{generated_samples_expectation}
\end{figure}

\noindent
\textbf{2} - for the second ablation experiment, the only modification of the architecture lies in the weights used for the different losses: all the introduced losses are used for the reconstruction output $D_R$, but for the entropic output $D_H$ only the encoder-conditioned image reconstruction loss remains. All other losses are removed. Samples of generated images for this study are available in Figure \ref{generated_samples_expectation}. The encoder-conditioned image reconstruction loss is not sufficient to ensure the complete removal of the shortcut, as can be seen with the samples of the Located-Patch CIFAR10 dataset.

\subsection{Experimental details for the large-scale experiments}

\noindent
\textbf{Synthetic Datasets}: the MNIST images are colorized and resized to 32x32 pixels, resulting in 3x32x32 pixel images, as the CIFAR10-based benchmarks images. For all synthetic experiments, the images are normalized with a mean and a standard deviation both of [0.5,0.5,0.5]. The validation set and the test set are both made with the images from the original MNIST or CIFAR10 test sets, but biased differently. There is no pretraining of the classifiers for these datasets.\\

\begin{figure*}[h!]
\centering
\includegraphics[width=17cm]{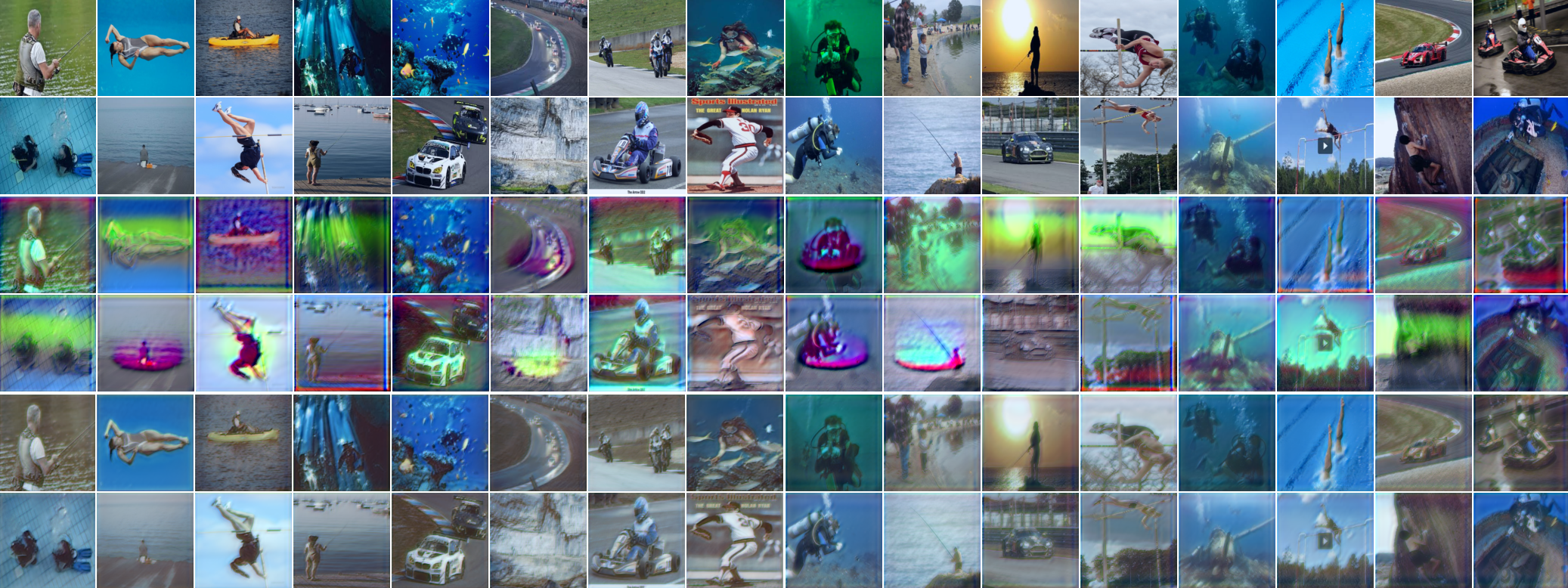}
\caption{Samples of original BAR images (first 2 rows) and their generated hybrid (middle 2 rows) and entropic (last 2 rows) versions. The entropic images are almost devoid of bright colors, showing that a classifier rely heavily on them instead on the causal patterns. The blue color is not removed as it is not strongly correlated with a particular class: diving, pole vaulting and fishing images exhibit large amount of blue. }
\label{generated_samples_bar}
\end{figure*}

\noindent
\textbf{BAR}: the BAR images are resized to 128x128 pixels, for computational convenience, and normalized with the following means and standard deviations: [0.485, 0.456, 0.406], and [0.229, 0.224, 0.225]. The validation set is made of 300 images that are removed from the original training set, keeping $\sim$ 1700 images for training. For all experiments on BAR, whether for our approach or existing methods, the ResNet18 is pretrained on ImageNet. Samples of the original BAR dataset can be found in Figure \ref{generated_samples_bar} alongside with their hybrid and entropic counterparts.\\

\noindent
\textbf{Standard Training Procedure}: for all datasets, we used a learning rate of $10^{-3}$ with the stochastic gradient descent (SGD) with a momentum of $0.9$ and trained for $100$ epochs, with a batch size of 128 for all datasets (all the experiments are trained with this batch size). The test-time model is selected by best accuracy on the validation set. In the case where the best accuracy is reached several time during training (a common phenomenon since it is easy to reach $100\%$ accuracy on the biased datasets), the model retained is the first to reach it.\\

\noindent
\textbf{Jigsaw}: images are divided into a 2x2 grid, whose patches are then shuffled. The weight for the permutation classification loss is fixed to 1.0 for BAR, 1.0 for the synthetic datasets. Early stopping was applied as soon as the validation accuracy reached $99\%$. Subsequently, the model selected was the last one. Other training hyper-parameters are the same as for the standard training procedure and this will also be the case for the next experiments, unless specified otherwise.\\

\noindent
\textbf{Dropout}: dropout was applied at the end of the features extractor part of the classifiers. It is before the last fully-connected layer for the ResNet18, and after the 2 convolutional layers for the LeNet. The zeroing probability is chosen randomly at each iteration.\\

\noindent
\textbf{Dropout \& Covariance Constraint}: the penalty is calculated with the $L_2$ norm of the covariance matrix of the features extractor activations computed over the current batch. The diagonal of the covariance matrix is fixed to 0 beforehand. The weight for the covariance penalty was fixed to $10^{-4}$ for all experiments.\\

\noindent
\textbf{Dropout \& Orthogonality Constraint}: the orthogonality constraint is calculated for all the layers, and the final constraint used is the average penalty over all layers. A particular layer's penalty is the $L_2$ norm of the dot product between a layer's weights and its transpose. The weight of the penalty is fixed to 1.0 for all experiments. Overall, the effects of these additional constraints compared to a simple dropout are negligible. Nonetheless, due to the apparent simplistic nature of our synthetic datasets, we deemed necessary to try naive approaches first to ensure that a more complex one was indeed needed to reach satisfying results.\\

\noindent
\textbf{RSC}: we used channel-wise dropout with a batch percentage of $100\%$ and the amount of channels dropped is randomly chosen at every step. Channels are sorted with respect to their usefulness for the classification on each samples in the batch, and a varying number of the most effective ones are dropped.\\

\noindent
\textbf{Spectral Decoupling}: the weight for the $L_2$ on the raw logits of the network (before the softmax) is fixed to 0.01 for BAR, and to 1.0 for the synthetic benchmarks. Early stopping is applied when the validation accuracy reaches $99\%$. \\

\noindent
\textbf{LfF}: the architecture was trained with Adam and a learning rate of $10^{-4}$, for $100$ epochs, with an amplification factor $q = 0.7$, for all the experiments.\\

\noindent
\textbf{LDD}: the architecture was trained with Adam and a learning rate of $10^{-4}$, for 100 epochs. The amplification factor is the same as for LfF. The hyper-parameters specific to LDD ($\lambda_{dis}, \lambda_{swap_b}, \lambda_{swap}$) were kept to the default value: [1.0,1.0,1.0], used in the original paper for datasets similar to ours. The bias-conflicting augmentation is scheduled to be applied after the first epoch for BAR and after the default value (10k iterations) for the synthetic datasets. Without counter-examples, this parameter has no impact on the results. \\

\noindent
\textbf{JTT}: one of the core principle in JTT is to train the first network only for a limited amount of epochs, to avoid overfitting on the train set and keep a few number of misclassified train samples. For our synthetic experiments, the perfect classifier was reached before the end of the first epoch. To properly adapt the method we stopped the training after a $99\%$ accuracy on the current batch was reached for the 10-th time since the beginning. On BAR, the first network was trained for $1$ epoch, before switching to training on the over-sampled dataset.\\

\noindent
\textbf{Ours}: The architecture was trained for $100$ epochs, with Adam and a learning rate of $10^{-4}$. The hyper-parameters used were: $\alpha=1.0$, $\beta=\gamma=0.1$, $\varepsilon=10^{-3}$, $\mu=1.0$, $\nu=0.1$, for both the experiments on the synthetic datasets and BAR.

\subsection{Small-scale experiments}

To demonstrate the wide range of applicability of our method, we experiment with a small-scale version of our architecture. The classifier used is a LeNet (taken from \cite{carlucci2019domain}), the encoder (respectively the decoder) is a custom network with 4 convolutional (respectively transposed convolutional) layers. The architecture details are available in Table \ref{encoder_decoder}. The decoder has 256 input channels while the encoder only has 128 output channels, because the LeNet features extractor (layers parts of the features extractor are marked in bold) output also has 128 channels. There are no skip-connections, and the encoder and decoder are single-output and single-input. The hyper-parameters used for the small-scale experiments are exactly the same as before for our approach and the debiasing methods. Some domain generalization algorithms however required different hyper-parameters: they were only trained for $20$ epochs, the weight for the logits norm minimization in Spectral Decoupling was fixed to 5.0, the weight for the permutation classification loss in Jigsaw was fixed to 10.0 and the covariance norm minimization weight used was $10^{-3}$. The results of our small-scale architecture, alongside with the debiasing and domain generalization methods (also evaluated with our LeNet), are available in Table \ref{sup_mat_results_low_scale}.\\

\begin{table}[h!]
\scriptsize
\caption{Small-scale networks architecture}
\label{encoder_decoder}
\centering
\begin{tabular}{|C{2.1cm}||C{2.1cm}||c|}
\hline
\multicolumn{3}{|c|}{Syntax follows PyTorch format} \\
\hline
LeNet & Encoder & Decoder\\
\hline\hline
\textbf{Conv2d(3,64,5)} & Conv2d(3,32,3) & ConvTranspose2d(256,64,4,2)\\
\textbf{ReLU} & ReLU & ReLU \\
\textbf{MaxPool2d(2,2)} & Conv2d(32,64,3) & ConvTranspose2d(64,64,3)\\
\textbf{Conv2d(64,128,5)} & ReLU & ReLU \\
\textbf{ReLU} & Conv2d(64,64,3) & ConvTranspose2d(64,32,3)\\
\textbf{MaxPool2d(2,2)} & ReLU & ReLU\\
Linear(3200, 1024) & Conv2d(64,128,3,2) & ConvTranspose2d(32,6,3)\\
ReLU &  ReLU & \\
Linear(1024,1024) & & \\
ReLU & & \\
Linear(1024, 10) & & \\
\hline
\end{tabular}
\end{table}

Our method is still effective, but the difference between the results on the original datasets and ours is shortened compared to the large-scale version. The debiasing methods perform as bad as with the large-scale version, which was to be expected again. Interestingly, the small-scale version of the domain generalization methods perform better than their large-scale counterparts, especially on the Colored-MNIST dataset. While for the large-scale setting all of these performed similarly on all 3 benchmarks, there are clear differences in the small-scale setting: most of the strategies now perform better on the Colored-MNIST benchmark. Dropout-based methods and Jigsaw yield an important increase of accuracy on the Colored-MNIST, but not on the CIFAR10-based benchmarks. Notably, Spectral Decoupling produces an increase of accuracy in all 3 benchmarks, even if not the best on Colored-MNIST. \\ 

To compare our method with the debiasing strategies in a situation where they should be able to perform well, we design a version of our benchmarks with counter-examples. For $1\%$ of the samples, the shortcut \textit{e.g.} the red patch position, is chosen randomly (with equal probability for each shortcuts) and not based on the image label. The test set is the same as the original one, with an average shortcut applied to each image. The small-scale results of these experiments are available in Table \ref{sup_mat_results_low_scale_ce}. The hyper-parameters used are the same as the ones used for the previous small-scale experiments, except for the starting bias-conflicting augmentation iteration in LDD. In a situation with counter-examples, the scheduling is important. We start the bias-conflicting augmentation after the first training epoch for all datasets.\\

The difference of behaviour between Colored-MNIST and the other benchmarks has widened compared to the situation without counter-examples: standard training reaches satisfying accuracy on Colored-MNIST, while having no positive impact on the others. Likewise, domain generalization methods only yield good results on the Colored-MNIST. As expected, the behaviour of debiasing methods changes completely. They all perform almost perfectly on the MNIST-based benchmark, and produce a noticeable accuracy increase on the others benchmarks. LDD's performance are, on average, on par with our method, except in the Located-CIFAR10 dataset where it reaches better test accuracy at the cost of a large drop in validation accuracy. Our small-scale experiments show the necessity of trying the various methods on diverse benchmarks, as the strength of the correlation between the shortcuts and the labels is not the only factor of success for a method.

\begin{table*}[t!]
\caption{Small-scale experiments}
\label{sup_mat_results_low_scale}
\begin{center}
\begin{tabular}{|l||c|c|c|c|c|c| V|}
\hline
Dataset $\rightarrow$ & \multicolumn{2}{c|}{Colored-MNIST} & \multicolumn{2}{c|}{Colored-Patch CIFAR10} & \multicolumn{2}{c|}{Located-Patch CIFAR10}\\
\hline
Method $\downarrow$ & Val. Acc. & Test Acc. & Val. Acc. & Test Acc. & Val. Acc. & Test Acc.\\
\hline\hline
Standard Training Procedure & $100 \pm 0.0$ &$27.2 \pm 3.3$ & $100 \pm 0.0$ & $13.2 \pm 3.7$ & $100 \pm 0.0$ & $15.5 \pm 0.5$ \\
\hline
Dropout & $100 \pm 0.0$ & $43.0 \pm 5.0$ & $100 \pm 0.0$ & $15.8 \pm 3.5$ & $100 \pm 0.0$ & $15.2 \pm 0.8$ \\
\hline
Dropout \& Orthogonality \cite{ortho} & $100 \pm 0.0$ & $36.5 \pm 1.7$ & $100 \pm 0.0$ & $15.6 \pm 1.5$ & $100 \pm 0.0$ & $24.0 \pm 0.5$\\
\hline
Dropout \& Covariance \cite{correlation} & $100 \pm 0.0$ & $34.9 \pm 3.9$ & $100 \pm 0.0$ & $14.8 \pm 2.0$ & $100 \pm 0.0$ & $20.9 \pm 1.8$\\
\hline
Jigsaw Puzzle \cite{carlucci2019domain} & $98.3 \pm 0.2$ & $65.9 \pm 4.8$ & $97.7 \pm 0.3$ & $22.0 \pm 1.0$ & $99.7 \pm 0.1$ & $21.3 \pm 0.4$
 \\
\hline
Spectral Decoupling \cite{pezeshki2020gradient} & $99.6 \pm 0.1$ & $49.1 \pm 2.5$ & $95.4 \pm 0.2$ & $30.5 \pm 1.0$ & $96.9 \pm 1.8$ & $29.1 \pm 1.2$\\
\hline
RSC \cite{huangRSC2020} & $99.7 \pm 0.0$ & $45.0 \pm 0.6$ & $95.8 \pm 2.0$ & $14.5 \pm 2.0$ & $100 \pm 0.0$ & $11.4 \pm 0.1$ \\
\hline
LfF \cite{nam2020learning} & $100 \pm 0.0$ & $23.9 \pm 5$ & $100 \pm 0.0$ & $14.8 \pm 1.8$ & $100 \pm 0.0$ & $15.3 \pm 2.1$ \\
\hline
JTT \cite{liu2021just} & $100 \pm 0.0$ & $29.2 \pm 6.7$ & $100 \pm 0.0$ & $15.4 \pm 2.8$ & $100 \pm 0.0$ & $16.1 \pm 0.4$\\
\hline
LDD \cite{lee2021learning} & $100 \pm 0.0$ & $12.2 \pm 2.7$ & $100 \pm 0.0$ & $14.2 \pm 4.1$ & $100 \pm 0.0$ & $10.0 \pm 0.0$ \\
\hline
\textbf{Ours} & $\mathbf{99.9 \pm 0.0}$ & $\mathbf{98.4 \pm 1.5}$ & $\mathbf{94.3 \pm 0.2}$ & $\mathbf{69.2 \pm 2.0}$ & $\mathbf{97.9 \pm 0.5}$ & $\mathbf{67.3 \pm 0.3}$\\
\hline
Standard Training Procedure & $99.2 \pm 0.04$ & $99.2 \pm 0.04$ & $72.8 \pm 0.4$ & $72.8 \pm 0.4$ & $72.8 \pm 0.4$ & $72.8 \pm 0.4$\\
on the original datasets & & & & & & \\
\hline
\end{tabular}
\end{center}
\end{table*}

\begin{table*}[t!]
\caption{Small-scale experiments on datasets with counter-examples}
\label{sup_mat_results_low_scale_ce}
\begin{center}
\begin{tabular}{|l||c|c|c|c|c|c| V|}
\hline
Dataset $\rightarrow$ & \multicolumn{2}{c|}{Colored-MNIST} & \multicolumn{2}{c|}{Colored-Patch CIFAR10} & \multicolumn{2}{c|}{Located-Patch CIFAR10}\\
\hline
Method $\downarrow$ & Val. Acc. & Test Acc. & Val. Acc. & Test Acc. & Val. Acc. & Test Acc.\\
\hline\hline
Standard Training Procedure & $99.4 \pm 0.1$ & $65.9 \pm 4.6$ & $99.1 \pm 0.0$ & $18.1 \pm 3.8$ & $99.2 \pm 0.0$ & $27.1 \pm 2.1$ \\
\hline
Dropout & $99.4 \pm 0.0$ & $67.3 \pm 6.1$ & $99.2 \pm 0.0$ & $17.3 \pm 4.9$ & $99.3 \pm 0.0$ & $22.9 \pm 2.9$ \\
\hline
Dropout \& Orthogonality \cite{ortho} & $99.4 \pm 0.0$ & $68.0 \pm 2.0$ & $99.0 \pm 0.1$ & $26.8 \pm 2.0$ & $99.1 \pm 0.1$ & $27.5 \pm 0.2$ \\
\hline
Dropout \& Covariance \cite{correlation} & $99.4 \pm 0.0$ & $59.6 \pm 5.0$ & $99.14 \pm 0.0$ & $22.9 \pm 1.6$ & $99.2 \pm 0.0$ & $17.0 \pm 5.4$\\
\hline
Jigsaw Puzzle \cite{carlucci2019domain} & $98.0 \pm 0.2$ & $72.9 \pm 0.9$ & $95.8 \pm 0.5$ & $23.6 \pm 0.6$ & $98.5 \pm 0.1$ & $20.8 \pm 1.8$ \\
\hline
Spectral Decoupling \cite{pezeshki2020gradient} & $98.8 \pm 0.2$ & $49.6 \pm 1.3$ & $95.3 \pm 0.2$ & $30.4 \pm 1.1$ & $96.9 \pm 1.1$ & $29.0 \pm 2.4$\\
\hline
RSC \cite{huangRSC2020} & $98.6 \pm 0.0$ & $72.8 \pm 5.5$ & $96.3 \pm 0.4$ & $17.7 \pm 2.3$ & $99.2 \pm 0.0$ & $14.9 \pm 0.8$ \\
\hline
LfF \cite{nam2020learning} & $98.7 \pm 0.4$ & $94.8 \pm 0.8$ & $89.8 \pm 0.8$ & $37.73 \pm 0.2$ & $87.3 \pm 3.1$ & $40.4 \pm 1.3$ \\
\hline
JTT \cite{liu2021just} & $99.8 \pm 0.0$ & $94.1 \pm 0.1$ & $98.6 \pm 0.0$ & $30.5 \pm 0.7$ & $98.7 \pm 0.0$ & $40.7 \pm 0.5$ \\
\hline
LDD \cite{lee2021learning} & $99.5 \pm 0.1$ & $98.2 \pm 0.2$ & $86.1 \pm 1.0$ & $67.6 \pm 0.6$ & $86.5 \pm 1.2$ & $69.4 \pm 0.7$ \\
\hline
\textbf{Ours} & $\mathbf{99.8 \pm 0.1}$ & $\mathbf{97.6 \pm 0.1}$ & $\mathbf{88.0 \pm 0.2}$ & $\mathbf{68.9 \pm 0.2}$ & $\mathbf{95.9 \pm 0.3}$ & $\mathbf{64.7 \pm 0.3}$\\
\hline
Standard Training Procedure & $99.2 \pm 0.0$ & $99.2 \pm 0.0$ & $72.7 \pm 0.4$ & $72.7 \pm 0.4$ & $72.7 \pm 0.4$ &$72.7 \pm 0.4$ \\
on the original datasets & & & & & & \\
\hline
\end{tabular}
\end{center}
\end{table*}

\end{document}